\documentclass[runningheads]{llncs}

\usepackage{times}  
\usepackage{helvet}  
\usepackage{courier}  
\usepackage{url}  
\usepackage{graphicx}  

\usepackage{epsfig}
\usepackage{amsmath}

\usepackage{amssymb}
\usepackage{epsfig}
\usepackage{url}
\usepackage{algorithm}
\usepackage{mathtools}

\usepackage{multirow}
\usepackage{booktabs}
\usepackage{balance}

\usepackage{subfig}

\newcommand{\norm}[1]{\left\lVert#1\right\rVert}

\begin{document}

\title{A Unified Framework for Domain Adaptation using Metric Learning on Manifolds}

\author{Sridhar Mahadevan\inst{1} \and
Bamdev Mishra \inst{2} \and
Shalini Ghosh \inst{3}}
\authorrunning{S. Mahadevan et al.}
%
\institute{University of Massachusetts. Amherst, MA 01003, and Department of Computer Science, Stanford University, Palo Alto, CA 94305 \\
\email{mahadeva@cs.umass.edu} \and
Microsoft.com, Hyderabad 500032, Telangana, India\\
\email{bamdevm@microsoft.com}\\
 \and
SRI International, Menlo Park, CA 94025\\
\email{shalini.ghosh@sri.com}}

\maketitle

\begin{abstract}
We present a novel framework for domain adaptation, whereby both {\em geometric} and {\em statistical} differences between a labeled source domain and unlabeled target domain can be reconciled using a unified mathematical framework that exploits the curved Riemannian geometry of statistical manifolds. It is assumed the feature distribution across the source and target domains are sufficiently dissimilar so that a classifier trained on the source domain would not perform well on the target domain (e.g., ranking Amazon book reviews given labeled movie reviews). Various approaches to domain adaptation have been studied in the literature, ranging from geometric approaches to statistical approaches. Our approach is based on formulating transfer from source to target as a problem of geometric mean metric learning on manifolds. Specifically, we exploit the curved Riemannian manifold geometry of symmetric positive definite (SPD) covariance matrices. We exploit a simple but important observation that as the space of covariance matrices is both a Riemannian space as well as a homogeneous space, the shortest path geodesic between two covariances on the manifold can be computed analytically. Statistics on the SPD matrix manifold, such as the geometric mean of two SPD matrices can be reduced to solving the well-known Riccati equation. We show how the Riccati solution can be constrained to not only reduce the statistical differences between the source and target domains, such as aligning second order covariances and minimizing the maximum mean discrepancy,  but also the underlying geometry of the source and target domains using diffusions on the underlying source and target manifolds. A key strength of our proposed approach is that it enables integrating multiple sources of variation between source and target in a unified way, by reducing the combined objective function to a nested set of Riccati equations where the solution can be represented by a cascaded series of geometric mean computations. In addition to showing the theoretical optimality of our solution, we present detailed experiments using standard transfer learning testbeds from computer vision comparing our proposed algorithms to past work in domain adaptation, showing improved results over a large variety of previous methods.

\end{abstract}



\section{Introduction}
When we apply machine learning \cite{murphy2013machine} to real-world problems, e.g., in image recognition~\cite{cacm:imagenet} or speech recognition~\cite{speech-dl}, a significant challenge is the need for having large amounts of (labeled) training data, which may not always be available. Consequently, there has been longstanding interest in developing machine learning techniques that can transfer knowledge across domains, thereby alleviating to some extent the need for training data as well as the time required to train the machine learning system. A detailed survey of transfer learning is given in \cite{tl:pan}.

Traditional machine learning assumes that the distribution of test examples follows that of the training examples \cite{pac}, whereas in transfer learning, this assumption is usually violated. {\em Domain adaptation} (DA) is a well-studied formulation of transfer learning that is based on developing methods that deal with the change of distribution in test instances as compared with training instances \cite{da:nips,daume:2009}. In this paper, we propose a new framework for domain adaptation, based on formulating transfer from source to target as a problem of geometric mean metric learning on manifolds. Our proposed approach enables integrating multiple sources of variation between source and target in a unified framework with a theoretically optimal solution. We also present detailed experiments using standard transfer learning testbeds from computer vision, showing how our proposed algorithms give improved results compared to existing methods. \\

\noindent {\bf Background:} One standard approach of domain adaptation is based on modeling the {\em covariate shift} \cite{cov-shift}.  Unlike traditional machine learning, in DA, the training and test examples are assumed to have different distributions.   It is usual in DA to categorize the problem into different types: (i) semi-supervised domain adaptation (ii)  unsupervised domain adaptation (iii) multi-source domain adaptation (iv) heterogeneous domain adaptation. 

Another popular approach to domain adaptation is based on aligning the distributions between source and target domains. A common strategy is based on the maximum mean discrepancy (MMD) metric \cite{mmd}, which is a nonparametric technique for measuring the dissimilarity of empirical distributions between source and target domains. Domain-invariant projection is one method that seeks to minimize the MMD measure using optimization on the Grassmannian manifold of fixed-dimensional subspaces of $n$-dimensional Euclidean space \cite{dip}. 

Linear approaches to domain adaptation involve the use of alignment of lower-dimensional subspaces or covariances from a data source domain $D_s = \{ x^s_i \}$ with labels $L_s = \{ y_i \}$  to a target data domain $D_t = \{ x^t_i \}$. We assume both $x^s_i$ and $x^t_i$ are $n$-dimensional Euclidean vectors, representing the values of $n$ features of each training example. One popular approach to domain adaptation relies on first projecting the data from the source and target domains onto a low-dimensional subspace, and then finding correspondences between the source and target subspaces. Of these approaches, the most widely used one is Canonical Correlation Analysis (CCA) \cite{hotelling}, a standard statistical technique used in many applications of machine learning and bioinformatics.  Several nonlinear versions \cite{kcca} and deep learning variants \cite{deep-cca} of CCA have been proposed. These methods often require explicit correspondences between the source and target domains to learn a common subspace. Because CCA finds a linear subspace, a family of manifold alignment methods have been developed that extend CCA \cite{unsup-ma,genunsup-ma} to exploit the nonlinear structure present in many datasets. 

In contrast to using a single shared subspace across source and target domains, {\em subspace alignment} finds a linear mapping that transforms the source data subspace into the target data subspace \cite{sa}. To explain the basic algorithm, let  $P_S, P_T \in \mathbb{R}^{n \times d}$ denote the two sets of basis vectors that span the subspaces for the ``source" and `` target" domains. Subspace alignment attempts to find a linear mapping $M$ that minimizes  
\begin{equation*}
\label{sa}
F(M) = \norm{ P_S M - P_T }^2_F.
\end{equation*} 
It can be shown that the solution to the above optimization problem is simply the dot product between $P_S$ and $P_T$, i.e.,:
\begin{equation*} 
\label{sa2} 
M^* = \text{argmin}_M F(M) = P_S^T P_T.
\end{equation*} 

Another approach exploits the property that the set of $k$-dimensional subspaces in $n$-dimensional Euclidean space forms a curved manifold called the {\em Grassmannian} \cite{edelman-siam}, a type of matrix manifold. The domain adaptation method called geodesic flow kernels (GFK) \cite{gfk} is based on constructing a distance function between source and target subspaces that is based on the geodesic or shortest path between these two elements on the Grassmannian.

Rather than aligning subspaces, a popular technique called CORAL \cite{coral} aligns correlations between source and target domains. Let $\mu_s, \mu_t$ and $A_s, A_t$ represent the mean and covariance of the source and target domains, respectively. CORAL finds a linear transformation $A$ that minimizes the distance between the second-order statistics of the source and target features (which can be assumed as normalized with zero means). Using the Frobenius (Euclidean) norm as the matrix distance metric, CORAL is based on solving the following optimization problem: 
\begin{equation} 
\label{coral}
\min_A \norm{ A^T A_s A - A_t }^2_F.
\end{equation} 
where $A_s, A_t$ are of size $n\times n$. Using the singular value decomposition of $A_s$ and $A_t$, CORAL \cite{coral} computes \emph{a} particular closed-form solution \footnote{ The solution characterization in \cite{coral} is non unique. \cite[Theorem~1]{coral} shows that the optimal $A$, for full-rank $A_s$ and $A_t$, is characterized as $A = U_s \Sigma_s^{-1} U_s^T U_t \Sigma_t^{1/2}U_t^T$, where $ U_s \Sigma_s U_s^T$ and $U_t \Sigma_tU_t^T$ are the eigenvalue decompositions of $A_s$ and $A_t$, respectively. However, it can be readily checked that there exists a continuous set of optimal solutions characterized as $A = U_s \Sigma_s^{-1/2}  U_s^T OU_t \Sigma_t^{1/2}U_t^T$, where $O$ is any orthogonal matrix, i.e., $OO^T = O^TO = I$ of size $n\times n$. A similar construction for non-uniqueness of the CORAL solution also holds for rank deficient $A_s$ and $A_t$.} to find the desired linear transformation $A$. \\

\noindent {\bf Novelty of our Approach:} Our proposed solution differs from the above previous approaches in several fundamental ways: one, we explicitly model the space of covariance matrices as a curved Riemannian manifold of symmetric positive definite (SPD) matrices. Note the difference of two SPD matrices is not an SPD matrix, and hence they do not form a vector space. Second, our approach can be shown to be both unique and globally optimal, unlike some of the above approaches. Uniqueness and optimality derive from the fact that we reduce all domain adaptation computations to nested equations involving solving the well-known \emph{Riccati} equation \cite{bhatia:spd}.  

The organization of the paper is as follows.  In Section \ref{sec:domain_adaptation_metric_learning}, we show the connection between the domain adaptation problem to the metric learning problem. In particular, we discuss the Riccati point of view for the domain adaptation problem. Section \ref{sec:Riemannian_geometry} discusses briefly the Riemannian geometry of the space of SPD matrices. Sections \ref{sec:statistical_alignment} and \ref{sec:geometrical_diffusion} discuss additional domain adaptation formulations. Our proposed algorithms are presented in Sections \ref{sec:cascaded_weighted} and \ref{sec:algorithms}. Finally, in Section \ref{sec:experiments} we show the experimental results on the standard Office and the extended Office-Caltech10 datasets, where our algorithms show clear improvements over CORAL.

%

\section{Domain Adaptation using Metric Learning}\label{sec:domain_adaptation_metric_learning}

In this section, we will describe the central idea of this paper: modeling the problem of domain adaptation as a geometric mean metric learning problem. Before explaining the specific approach, it will be useful to introduce some background. The metric learning problem \cite{synth:metric} involves taking input data in $\mathbb{R}^n$ and constructing a (non)linear mapping $\Phi: \mathbb{R}^n \rightarrow \mathbb{R}^m$, so that the distance between two points $x$ and $y$ in $\mathbb{R}^n$ can be measured using the distance $\| \Phi(x) - \Phi(y) \|$. A simple approach is to learn a squared {\em Mahalanobis} distance: $\delta_A^2(x, y) = (x - y)^T A (x - y)$, where $x, y \in \mathbb{R}^n$ and $A$ is an $n \times n$ symmetric positive definite (SPD) matrix. If we represent $A = W^T W$, for some linear transformation matrix $W$, then it is easy to see that $\delta_A^2(x, y) = \| W x - W y\|_F^2$, thereby showing that the Mahalanobis distance is tantamount to projecting the data into a potentially lower-dimensional space, and measuring distances using Euclidean (Frobenius) norm. Typically, the matrix $A$ is learned using some weak supervision, given two sets of training examples of the form: 
\[ {\cal S} = \{ (x_i, x_j) | x_i \text{ and } x_j \text{ are  in the same class}  \},  \]
\[ {\cal D} = \{ (x_i, x_j) | x_i \text{ and } x_j \text{ are in different classes} \}.  \]
A large variety of metric learning methods can be designed based on formulating different optimization objectives based on functions over the $S$ and $D$ sets to extract information about the distance matrix $A$. 

For our purposes, the method that will provide the closest inspiration to our goal of designing a domain adaptation method based on metric learning is the recently proposed \emph{geometric mean metric learning} (GMML) algorithm \cite{gmml}. GMML models the distance between points in the $S$ set by the Mahalanobis distance $\delta_A(x_i, x_j), x_i, x_j \in S$ by exploiting the geometry of the SPD matrices, and crucially, also models the distance between points in the disagreement set $D$ by the \emph{inverse} metric $\delta_{A^{-1}}(x_i, x_j), x_i, x_j \in D$. GMML is based on solving the objective function over all SPD matrices $A$:
\begin {equation*}
\label{metric}
\min_{A \succ 0} \sum_{(x_i, x_j) \in S} \delta^2_A(x_i, x_j)  + \sum_{(x_i, x_j) \in D} \delta^2_{A^{-1}}(x_i, x_j), 
\end{equation*} 
where $A \succ 0$ refers to the set of all SPD matrices.

Several researchers have previously explored the connection between domain adaptation and metric learning. One recent approach is based on constructing a transformation matrix $A$ that both minimizes the difference between the source and target distributions based on the previously noted MMD metric, but also captures the manifold geometry of source and target domains, and attempts to preserve the discriminative power of the label information in the source domain \cite{wang-aaai14}. Our approach builds on these ideas, with some significant differences. One, we use an objective function that is based on finding a solution that lies on the geodesic between source and target (estimated) covariance matrices (which are modeled as symmetric positive definite matrices). Second, we use a cascaded series of geometric mean computations to balance multiple factors. We describe these ideas in more detail in this and the next section. 

We now describe how the problem of domain adaptation can be considered as a type of metric learning problem, called geometric mean metric learning (GMML) \cite{gmml}. Recall that in domain adaptation, we are given a source dataset $D_s$ (usually with a set of training labels) and a target dataset $D_t$ (unlabeled). The aim of domain adaptation, as reviewed above, is to construct an intermediate representation that combines some of the features of both the source and target domains, with the rationale being that the distribution of target features differs from that of the source. Relying purely on either the source or the target features is therefore suboptimal, and the challenge is to determine what intermediate representation will provide optimal transfer between the domains.

To connect metric learning to domain adaptation, note that we can define the two sets $S$ and $D$ in the metric learning problem as associated with the source and target domains respectively, whereby \footnote{We note that while there are alternative ways to define the $S$ and $D$ sets, the essence of our approach remains similar.}
\begin{eqnarray*}
S &\subseteq& D_s \times D_s =  \{ (x_i, x_j) | x_i \in D_s, x_j \in D_s \} \\
D &\subseteq& D_t \times D_t = \{ (x_i, x_j) | x_i \in D_t, x_j \in D_t \}. 
\end{eqnarray*} 
Our approach seeks to exploit the nonlinear geometry of covariance matrices to find a Mahalanobis distance matrix $A$, such that we can represent distances in the source domain using $A$, but crucially we measure distances in the target domain using the inverse $A^{-1}$. 
\begin {eqnarray*}
\label{gca-eqn}
\min_{A \succ 0} \sum_{(x_i, x_j) \in S} (x_i - x_j)^T A  (x_i - x_j)  
+ \sum_{(x_i, x_j) \in D} (x_i - x_j)^T A^{-1} (x_i - x_j).
\end{eqnarray*} 
To provide some intuition here, we observe that as we vary $A$ to reduce the distance $\sum_{(x_i, x_j) \in S} (x_i - x_j)^T A  (x_i - x_j) $ in the source domain, we simultaneously increase the distance in the target domain by minimizing $\sum_{(x_i, x_j) \in D} (x_i - x_j)^T A^{-1} (x_i - x_j)$, and vice versa. Consequently, by appropriately choosing $A$, we can seek the minimize the above sum. We can now use the matrix trace to reformulate the Mahalanobis distances: 
\begin {eqnarray*}
\label{gca-eqn2}
\min_{A \succ 0} \sum_{(x_i, x_j) \in S} {\rm tr}(A (x_i - x_j) (x_i - x_j)^T )  + \sum_{(x_i, x_j) \in D} {\rm tr}(A^{-1} (x_i - x_j) (x_i - x_j)^T ).
\end{eqnarray*} 
Denoting the source and target covariance matrices $A_s$ and $A_t$ as: 
\begin{eqnarray}
\label{covm}
A_s &\coloneqq& \sum_{(x_i, x_j) \in S} (x_i - x_j) (x_i - x_j)^T \\
\label{covm-2}
A_t &\coloneqq&  \sum_{(x_i, x_j) \in D} (x_i - x_j) (x_i - x_j)^T,
\end{eqnarray}
we can finally write a new formulation of the domain adaptation problem as minimizing the following objective function to find the SPD matrix $A$ such that: 
\begin{equation}
\label{gca-eqn3}
\min_{A \succ 0}  \ \omega(A) \coloneqq {\rm tr}(A A_s) + {\rm tr}(A^{-1} A_t) .
\end{equation}

\section{Riemannian Geometry of SPD Matrices} \label{sec:Riemannian_geometry}

In this section, we outline some other formulations of domain adaptation that will be useful to discuss for presenting our overall approach.

\begin{figure}
	\centering
	\includegraphics[width=0.8\linewidth]{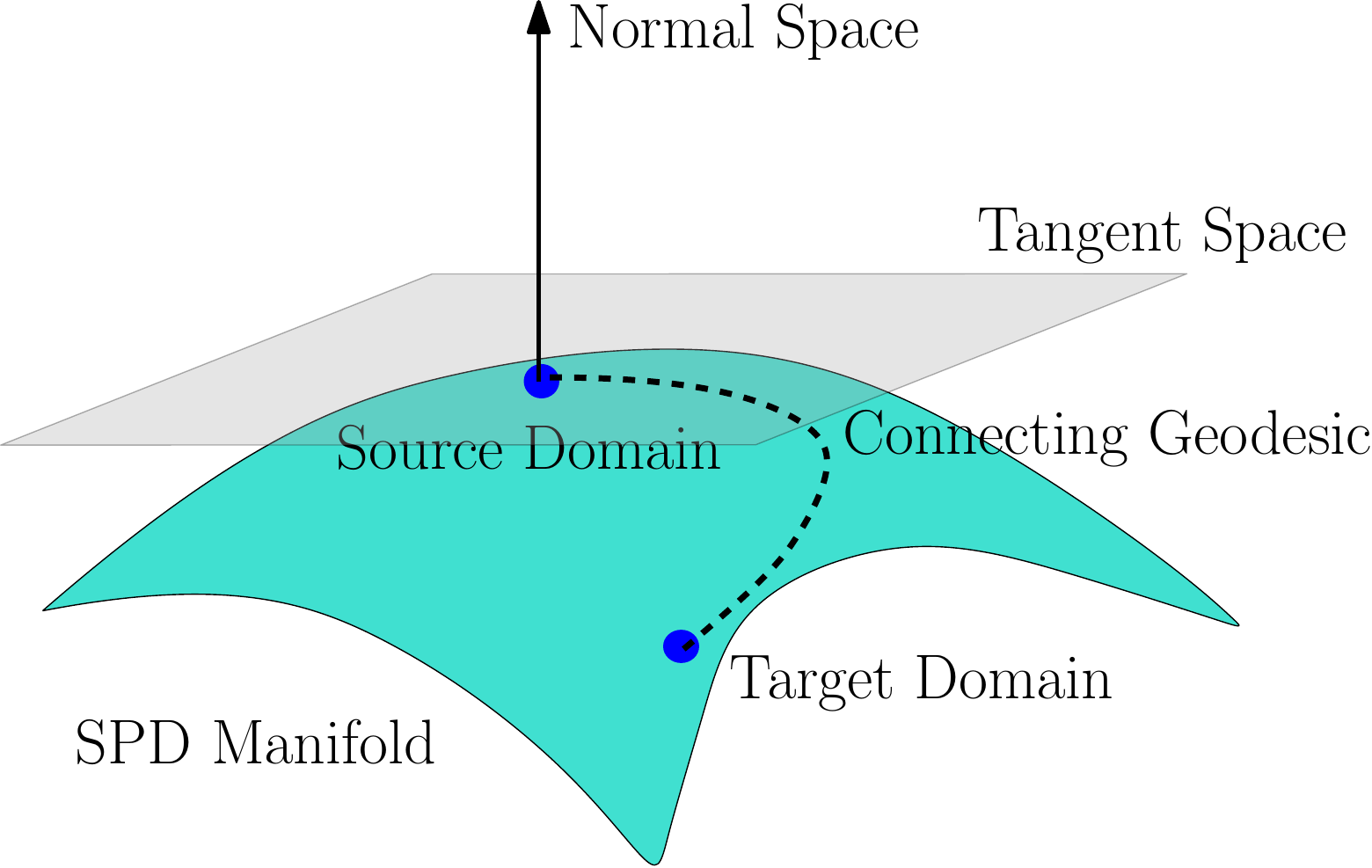}
	\caption{The space of symmetric positive definite matrices forms a Riemannian manifold, as illustrated here. The methods we propose are based on computing geodesics (the shortest distance), shown in the dotted line, between source domain information and target domain information. }
	\label{fig:psd}
\end{figure}

As Figure~\ref{fig:psd} shows, our proposed approach to domain adaptation builds on the nonlinear geometry of the space of SPD (or covariance) matrices, we review some of this material first \cite{bhatia:spd}. Taking a simple example of a $2 \times 2$ SPD matrix $M$, where: 
\begin{equation*}
M = \left[ \begin{array}{cc} a\quad & \quad b \\ b\quad  & \quad c \end{array} \right],
\end{equation*} 
where $a > 0$, and the SPD requirement implies the positivity of the determinant $ac - b^2 > 0$. Thus, the set of all SPD matrices of size $2 \times 2$ forms the interior of a cone in $\mathbb{R}^3$. More generally, the space of all $n \times n$ SPD matrices forms a manifold of non-positive curvature in $\mathbb{R}^{n^2}$ \cite{bhatia:spd}.
In the CORAL objective function in Equation~(\ref{coral}), the goal is to find a transformation $A$ that makes the source covariance resemble the target as closely as possible. Our approach simplifies Equation~(\ref{coral}) by restricting the transformation matrix $A$ to be a SPD matrix, i.e, $A \succ 0$, and furthermore, we solve the resulting nonlinear equation {\em exactly} on the manifold of SPD matrices. More formally, we solve the {\em Riccati equation} \cite{bhatia:spd}:
\begin{eqnarray}
\label{riccati1} 
 A A_s A = A_t, \text{  for  } A \succ 0,
\end{eqnarray} 
where $A_s$ and $A_t$ are source and target covariances SPD matrices, respectively. Note that in comparison with the CORAL approach in Equation~(\ref{coral}), the $A$ matrix is symmetric (and positive definite), so $A$ and $A^T$ are the same. The solution to the above Riccati equation is the well-known {\em geometric mean} or {\em sharp mean}, of the two SPD matrices, $A_s^{-1}$ and $A_t$. 
\begin{equation*}
\label{riccati}
A = A_s^{-1} \sharp_{\frac{1}{2}} A_t = A_s^{-\frac{1}{2}} (A_s^\frac{1}{2} A_t A_s^{\frac{1}{2}})^\frac{1}{2} A_s^{-\frac{1}{2}},
\end{equation*}
where $\sharp_{\frac{1}{2}}$ is denotes the geometric mean of SPD matrices \cite{gmml}. The sharp mean has an intuitive geometric interpretation: it is the midpoint of the geodesic connecting the source domain $A_s^{-1}$ and target domain $A_t$ matrices, where length is measured on the Riemannian manifold of SPD matrices. 

In a manifold, the shortest path between two elements, if it exists, can be represented by a geodesic. For the SPD manifold, it can be shown that the geodesic $\gamma(t)$ for a scalar $0 \leq t \leq 1$ between $A_s^{-1}$ and $A_t$, is given by \cite{bhatia:spd}: 
\begin{equation*}
\label{sharp-mean}
\gamma(t) = A_s^{-1/2} (A_s^{1/2} A_t A_s^{1/2})^t A_s^{-1/2}.
\end{equation*}
It is common to denote $\gamma(t)$ as the so-called ``weighted'' sharp mean $A_s^{-1} \sharp_t A_t$. It is easy to see that for $t=0$, $\gamma(0) = A_s^{-1}$, and for $t=1$, we have $\gamma(1) = A_t$. For the distinguished value of $t = 1/2$, it turns out that $\gamma(1/2)$ is the {\em geometric mean} of $A_s^{-1}$ and $A_t$, respectively, and satisfies all the properties of a geometric mean \cite{bhatia:spd}. 
\begin{equation*}
\label{geom-mean}
\gamma(1/2) = A_s^{-1/2} \sharp_{1/2} A_t = A_s^{-1/2} (A_s^{1/2} A_t A_s^{1/2})^{1/2} A_s^{-1/2}.
\end{equation*}
The following theorem summarizes some of the properties of the objective function given by Equation~(\ref{gca-eqn3}). 

\begin{theorem}\label{thm:geodesic_convexity}
\cite[Theorem~3]{gmml} The cost function $\omega(A)$ in Equation~(\ref{gca-eqn3}) is both strictly convex as well as strictly geodesically convex on the SPD manifold. 
\end{theorem}

Theorem \ref{thm:geodesic_convexity} ensures the uniqueness of the solution to Equation~(\ref{gca-eqn3}).

\section{Statistical Alignment Across Domains} \label{sec:statistical_alignment}

A key strength of our approach is that it can exploit both geometric and statistical information, and multiple sources of alignment are integrated by solving nested sets of Riccati equations. To illustrate this point, in this section we explicitly introduce a secondary criterion of aligning the source and target domains so that the underlying (marginal) distributions are similar. As our results show later, we obtain a significant improvement over CORAL on a standard computer vision dataset (Office/Caltech/Amazon problem). The reason our approach outperforms CORAL is that not only are we able to solve the Riccati equation uniquely, whereas the CORAL solution proposed  is only a particular solution due to non-uniqueness, whereas we can exploit multiple sources of information. 

A common way to incorporate the statistical alignment constraint is based on minimizing the maximum mean discrepancy metric (MMD) \cite{mmd}, a nonparametric measure of the difference between two distributions. 
\begin{equation} 
\label{mmd}
\norm{\frac{1}{n} \sum_{i=1}^n W x_i^s - \frac{1}{m} \sum_{i=1}^m W x_i^t }^2 = {\rm tr}(A X L X^T), 
\end{equation}
where $X = \{x_1^s, \ldots, x_n^s, x_t^1, \ldots, x_t^n\} \in \mathbb{R}^{(n+m)}$ and $L \in \mathbb{R}^{(n + m)(n+m)}$, where $L(i,j) = \frac{1}{n^2}$ if $x_i, x_j \in X_{s}$, $L(i,j) = \frac{1}{m^2}$ if $x_i, x_j \in X_t$, and $L(i,j) = -\frac{1}{mn}$ otherwise. It is straightforward to show that $L \succeq 0$, a symmetric positive-semidefinite matrix \cite{wang-aaai14}. 
We can now combine the MMD objective in Equation~(\ref{mmd}) with the previous geometric mean objective in Equation~(\ref{gca-eqn3}) to give rise to the following modified objective function:
\begin{equation}
\label{gca-mmd}
\min_{A \succ 0}  \ \xi(A) \coloneqq {\rm tr}(A A_s) + {\rm tr}(A^{-1} A_t) + {\rm tr}(A X L X^T).
\end{equation}
We can once again find a closed-form solution to the modified objective in Equation~(\ref{gca-mmd}) by taking gradients: 
\[ \nabla \xi(A) =  A_s - A^{-1} A_t A^{-1} + X L X^T = 0, \]
whose solution is now given $A = A_m^{-1} \sharp_{\frac{1}{2}} A_t$, where $A_m = A_s + X L X^T$.

%

\section{Geometrical Diffusion on Manifolds} \label{sec:geometrical_diffusion}

Thus far we have shown how the solution to the domain adaptation problem can be shown to involve finding the geometric mean of two terms, one involving the source covariance information and the Maximum Mean Discrepancy (MMD) of source and target training instances, and the second involving the target covariance matrix. In this section, we impose additional geometrical constraints on the solution that involve modeling the nonlinear manifold geometry of the source and target domains. 

The usual approach is to model the source and target domains as a nonlinear manifold and set up a diffusion on a discrete graph approximation of the continuous manifold \cite{diffmap}, using a random walk on a nearest neighbor graph connecting nearby points. Standard results have been established showing asymptotic convergence of the graph Laplacian to the underlying manifold Laplacian \cite{convle}. We can use the above algorithm to find two graph kernels $K_s$ and $K_t$ that are based on the eigenvectors of the random walk on the source and target domain manifold, respectively. 
\begin{eqnarray*}
\label{kernel}
K_s = \sum_{i=1}^m e^{-\frac{-\sigma^2_s}{2 \lambda^s_i}} v^s_i (v^s_i)^T \\
K_t = \sum_{i=1}^n e^{-\frac{-\sigma^2_t}{2 \lambda^t_i}} v^t_i (v^t_i)^T.
\end{eqnarray*}
Here, $v^s_i$ and $v^t_i$ refer to the eigenvectors of the random walk diffusion matrix on the source and target manifolds, respectively, and $\lambda^s_i$ and $\lambda^t_i$ refer to the corresponding eigenvalues.

We can now introduce a new objective function that incorporates the source and target domain manifold geometry:
\begin{equation}
\label{gfab-eqn} 
 \min_{A \succ 0} \ \eta(A) \coloneqq {\rm tr}(A X (K + \mu L) X^T) + {\rm tr}(A A_s) + {\rm tr}(A^{-1} A_t),
\end{equation} 
where $K \succ 0$ and $K = \left(\begin{array}{cc} K^{-1}_s & 0 \\ 0 & K^{-1}_t \end{array} \right)$, and $\mu$ is a weighting term that combines the geometric and statistical constraints over $A$.

Once again, we can exploit the SPD nature of the matrices involved, the closed-form solution to Equation (\ref{gfab-eqn}) is $A = A_{gs} \sharp_{\frac{1}{2}} A_t$, where $A_{gs} = A_s + X (K + \mu L) X^T$.


\section{Cascaded Weighted Geometric Mean} \label{sec:cascaded_weighted}
One additional refinement that we use is the notion of a {\em weighted} geometric mean. To explain this idea, we introduce the following Riemannian distance metric on the nonlinear manifold of SPD matrices: 
\begin{equation*}
\label{logm}
\delta^2_R(X,Y) \equiv \norm{ \log (Y^{-\frac{1}{2}} X Y^{-\frac{1}{2}}) }^2_F
\end{equation*}
%
%
for two SPD matrices $X$ and $Y$. Using this metric, we now introduce a {\em weighted} version of the previous objective functions in Equations (\ref{gca-eqn3}), (\ref{gca-mmd}), and (\ref{gfab-eqn}). 

For the first objective function in Equation~(\ref{gca-eqn3}), we get:
\begin{equation}
\label{gca-eqn-wt}
\min_{A \succ 0}\ \omega_t(A) \coloneqq (1-t) \ \delta_R^2(A, A_s^{-1}) + t \ \delta_R^2(A, A_t),
\end{equation}
where $0 \leq t \leq 1$ is the weight parameter. The unique solution to (\ref{gca-eqn-wt}) is given by the {\em weighted geometric mean} $A = A_s^{-1} \sharp_t A_t$ \cite{gmml}. Note that the weighted metric mean is no longer strictly convex (in the Euclidean sense), but remains geodesically strictly convex \cite[Chapter~6]{gmml,bhatia:spd}. 

Similarly, we introduce the weighted variant of the objective function given by Equation~(\ref{gca-mmd}): 
\begin{equation}
\label{gca-mmd-wt}
\min_{A \succ 0} \ \xi_t(A) \coloneqq  (1 -t) \ \delta^2_R(A, (A_s + XLX^T)^{-1}) + t \ \delta^2_R(A, A_t),
\end{equation}
whose unique solution is given by $A = A_{m}^{-1} \ \sharp_t \ A_t$, where $A_m = A_s + X L X^T$ as before. A cascaded variant is obtained when we further exploit the SPD structure of $A_s$ and $XLX^T$, i.e., $A_m = A_s \sharp_\gamma (X L X^T)$ (weighted geometric mean of $A_s$ and $(X L X^T$) instead of $A_m = A_s + X L X^T$ (which is akin to the Euclidean mean of $A_s$ and $(X L X^T$). Here, $0 \leq \gamma \leq 1$ is the weight parameter.

Finally, we obtain the weighted variant of the third objective function in Equation~(\ref{gfab-eqn}):
\begin{equation}
\label{gfab-eqn-wt}
\min_{A \succ 0} \ \eta_t(A) \coloneqq  (1-t)\delta_R^2(A, (A_s + X(K + \mu I)X^T)^{-1} + t \ \delta_R^2(A, A_t),
\end{equation}
whose unique solution is given by $A = A_{gs}^{-1} \ \sharp_t \ A_t$, where $A_{gs} = A_s + X (K + \mu L) X^T$ as previously noted. Additionally, the cascaded variant is obtained when $A_s \sharp_\gamma  (X (K + \mu L) X^T)$ instead of $A_{gs} = A_s + X (K + \mu L) X^T$.

\section{Domain Adaptation Algorithms} \label{sec:algorithms}

We now describe the proposed domain adaptation algorithms, based on the above development of  approaches reflecting geometric and statistical constraints on the inferred solution. All the proposed algorithms are summarized in Algorithm \ref{alg:gca-basic}. The algorithms are based on finding a Mahalanobis distance matrix $A$ interpolating source and target covariances (GCA1), incorporating an additional MMD metric (GCA2) and finally, incorporating the source and target manifold geometry (GCA3). It is noteworthy that all the variants rely on computation of the sharp mean, a unifying motif that ties together the various proposed methods. Modeling the Riemannian manifold underlying SDP matrices ensures the optimality and uniqueness of our proposed methods. 


\begin{algorithm}[ht]
\caption{Algorithms for Domain Adaptation using Metric Learning}
\label{alg:gca-basic}
	\small
	Given: A source dataset of labeled points $x^s_i \in D_s$  with labels $L_s = \{y_i \}$, and an unlabeled target dataset $x^t_i \in D_t$,  with hyperparameters $t, \mu$,  and $\gamma$. 
	\begin{enumerate}
		\item Define $X = \{x^s_1, \ldots, x^s_n, x^t_1, \ldots, x^t_m \}$.
		\item Compute the source and target matrices $A_s$ and $A_t$ using Equations~(\ref{covm}) and (\ref{covm-2}). 
		\item {\bf Algorithm GCA1:} Compute the weighted geometric mean $A = A^{-1}_s \ \sharp_t \ A_t$ (see Equation~(\ref{gca-eqn-wt})).
		\item {\bf Algorithm GCA2:} Compute the weighted geometric mean  taking additionally into account the MMD metric $A = A^{-1}_m \ \sharp_t \ A_t$ (see Equation~(\ref{gca-mmd-wt})), where $A_m = A_s + X L X^T$. 
		\item {\bf Algorithm Cascaded-GCA2:} Compute the \emph{cascaded} weighted geometric mean  taking additionally into account the MMD metric $A = A^{-1}_m \ \sharp_t \ A_t$ (see Equation~(\ref{gca-mmd-wt})), $A_m = A_s \sharp_\gamma X L X^T$. 
		\item {\bf Algorithm GCA3:} Compute the weighted geometric mean taking additionally into account the source and target manifold geometry $A = A^{-1}_{gs} \ \sharp_t \ A_t$ (see Equation~(\ref{gfab-eqn-wt})), where $A_{gs} = A_s + X (K + \mu L) X^T$.
		\item {\bf Algorithm Cascaded-GCA3:} Compute the \emph{cascaded} weighted geometric mean taking additionally into account the source and target manifold geometry $A = A^{-1}_{gs} \ \sharp_t \ A_t$ (see Equation~(\ref{gfab-eqn-wt})), where $A_{gs} = A_s \sharp_\gamma (X (K + \mu L) X^T)$.
		\item Use the learned $A$ matrix to adapt source features to the target domain, and perform classification (e.g., using support vector machines). 
	\end{enumerate}
\end{algorithm}

\section{Experimental Results}\label{sec:experiments}
We present experimental results using the standard computer vision testbed used in prior work: the Office \cite{gfk} and extended Office-Caltech10 \cite{sa} benchmark datasets. The Office-Caltech10 dataset contains $10$ object categories from an office environment (e.g., keyboard, laptop, and so on) in four image domains: Amazon ({\bf A}), Caltech256 ({\bf C}), DSLR ({\bf D}), and Webcam ({\bf W}). The Office dataset has $31$ categories (the previous $10$ categories and $21$ additional ones). 

An exhaustive comparison of the three proposed methods with a variety of previous methods is summarized by the table in Table~\ref{table1}. The previous methods compared in the table refer to the unsupervised domain adaptation approach where a support vector machine (SVM) classifier is used. The experiments follow the standard protocol established by previous works in domain adaptation using this dataset. The features used (SURF) are encoded with $800$-bin bag-of-words histograms and normalized to have zero mean and unit standard deviation in each dimension. As there are four domains, there are $12$ ensuing transfer learning problems, denoted in Table~\ref{table1} below as ${\bf A} \rightarrow {\bf D}$ (for Amazon to DSLR, etc.). For each of the 12 transfer learning tasks, the best performing method is indicated in boldface. We used $30$ randomized trials for each experiment, and randomly sample the same number of labeled images in the source domain as training set, and use all the unlabeled data in the target domain as the test set. All experiments used a support vector machine (SVM) method to measure classifier accuracy, using a standard {\tt libsvm} package. The methods compared against in Table~\ref{table1} include the following alternatives: 

\begin{itemize}
	\item {\bf Baseline-S}: This approach uses the projection defined by using PCA in the source domain to project both source and target data. 
	
	\item {\bf Baseline-T}: Here, PCA is used in the target domain to extract a low-dimensional subspace. 
	
	\item {\bf NA}: No adaptation is used and the original subspace is used for both source and target domains. 
	
	\item {\bf GFK}: This approach refers to the {\em geodesic flow kernel} \cite{gfk}, which computes the geodesic on the Grassmannian between the PCA-derived source and target subspaces computed from the source and target domains. 
	 
	\item {\bf TCA}: This approach refers to the transfer component analysis method \cite{tca}. 
	
	\item {\bf SA:} This approach refers to the subspace alignment method \cite{sa}. 
	
	\item {\bf CORAL:} This approach refers to the correlational alignment method \cite{coral}. 
	
	\item {\bf GCA1:} This is a new proposed method, based on finding the weighted geometric mean of the inverse of the source matrix $A_s$ and the target matrix $A_t$. 
	
	\item {\bf GCA2:} This is a new proposed method, based on finding the (non-cascaded) weighted geometric mean of the inverse of the source matrix $A_m$ and the target matrix $A_t$. 
			
	\item {\bf GCA3:} This is a new proposed method, based on finding the (non-cascaded) weighted geometric mean of the inverse of the source matrix $A_{gs}$ and the target matrix $A_t$. 
	
	\item {\bf Cascaded-GCA2:} This is a new proposed method, based on finding the cascaded geometric mean of the inverse revised source matrix $A_m$ and target matrix $A_t$ 
	
	\item {\bf Cascaded-GCA3:} This is a new proposed method, based on finding the cascaded geometric mean of the inverse revised source matrix $A_{gs}$  and target matrix $A_t$

\end{itemize}

\begin{table}[t]
\begin{minipage}[t]{1\textwidth}
\centering
\begin{tabular}{ccccccc}
\toprule
{\bf Method}  & {\bf C $\rightarrow$ A}  & {\bf D $\rightarrow$ A} & {\bf W $\rightarrow$ A} & {\bf A $\rightarrow$ C}  & {\bf D $\rightarrow$ C} & {\bf W $\rightarrow$ C} \\ 
\midrule
NA & 44.0 & 34.6 & 30.7 & 35.7 &  30.6 & 23.4 \\ 
B-S & 44.3 & 36.8 & 32.9 & 36.8 & 29.6 & 24.9 \\  
B-T  & 44.5 & 38.6 & 34.2 & 37.3 & 31.6 & 28.4 \\  
GFK  & 44.8 & 37.9 & 37.1 & 38.3 & 31.4 & 29.1 \\ 
TCA  & 47.2 & 38.8 & 34.8 & 40.8 & 33.8 & 30.9 \\  
SA  & 46.1 & {\bf 42.0} & {\bf 39.3} & 39.9 & 35.0 & 31.8 \\ 
CORAL  & 47.1 & 38.0 & 37.7 & 40.5 & 33.9 & 34.4 \\  
{\bf GCA1}  &  48.4 &  40.1 & 38.6  & {\bf 41.4} & {\bf 36.0} & {\bf 35.0 }\\ 
{\bf GCA2}  & {\bf 48.9 } &  40.0 & 38.6  &  41.0 & 35.9 & {\bf 35.0 }\\  
{\bf GCA3}  & 48.4 &  40.9 & 37.6  & 41.1 & 36 & 33.6\\  
{\bf Cascaded-GCA2}  & {\bf 49.5} &  41.1 & 38.6  &  41.1 & {\bf 36} & {\bf 35.1}\\  
{\bf Cascaded-GCA3} & {\bf 49.4} &  41.0 & 38.5  &  41.0 & 35.9 & 35 \\ 
\bottomrule
\\ 
\toprule
{\bf Method}  & {\bf A $\rightarrow$ D}  & {\bf C $\rightarrow$ D} & {\bf W $\rightarrow$ D} & {\bf A $\rightarrow$ W}  & {\bf C $\rightarrow$ W} & {\bf D $\rightarrow$ W} \\ 
\midrule
NA & 34.5 & 36.0 & 67.4 & 26.1 & 29.1 & 70.9 \\  
B-S & 36.1 & 38.9 & 73.6 & {\bf 42.5} & 34.6 & 75.4 \\  
B-T  & 32.5 & 35.3 & 73.6 & 37.3 & 34.2 & 80.5 \\ 
GFK  & 32.5 & 36.1 & 74.6 & 39.8 & 34.9 & 79.1 \\ 
TCA  & 36.4 & 39.2 & 72.1 & 38.1 & 36.5 & 80.3 \\  
SA  & {\bf 44.5} & 38.6 & 34.2 & 37.3 & 31.6 & 28.4 \\  
CORAL  & 38.1 & 39.2 & 84.4 & 38.2 & 39.7 & 85.4 \\  
{\bf GCA1 }  & 39.2 & 40.9 & 85.1 & 40.9 & 41.1 & {\bf 87.2}  \\ 
{\bf GCA2}  & 39.9  & 41.4  &  85.1  & 40.1 & 41.1 & {\bf 87.2 }\\ 
{\bf GCA3}  & 38.9 &   41.6  & 84.9 &  39.9 & {\bf 41.4} & 86.8\\
{\bf Cascaded-GCA2}  & 40.4 &  40.7 & {\bf 85.5}  &  41.3 & 40.3 & {\bf 87.2}\\ 
{\bf Cascaded-GCA3} & 38.8 &  {\bf 42.4}  & 85.3  &  40.1 & 40.9 & 86.9\\ 
\bottomrule  
\end{tabular}
\end{minipage} 
\caption{Recognition accuracy with unsupervised domain adaptation using SVM classifier (Office dataset + Caltech10). Our proposed methods (labeled GCAXX and Cascaded-GCAXX) perform better than all other methods in a majority of the 12 transfer learning tasks.} 
\label{table1} 
\end{table}

\begin{figure}[p]
	\centering
	\includegraphics[width=0.8\linewidth]{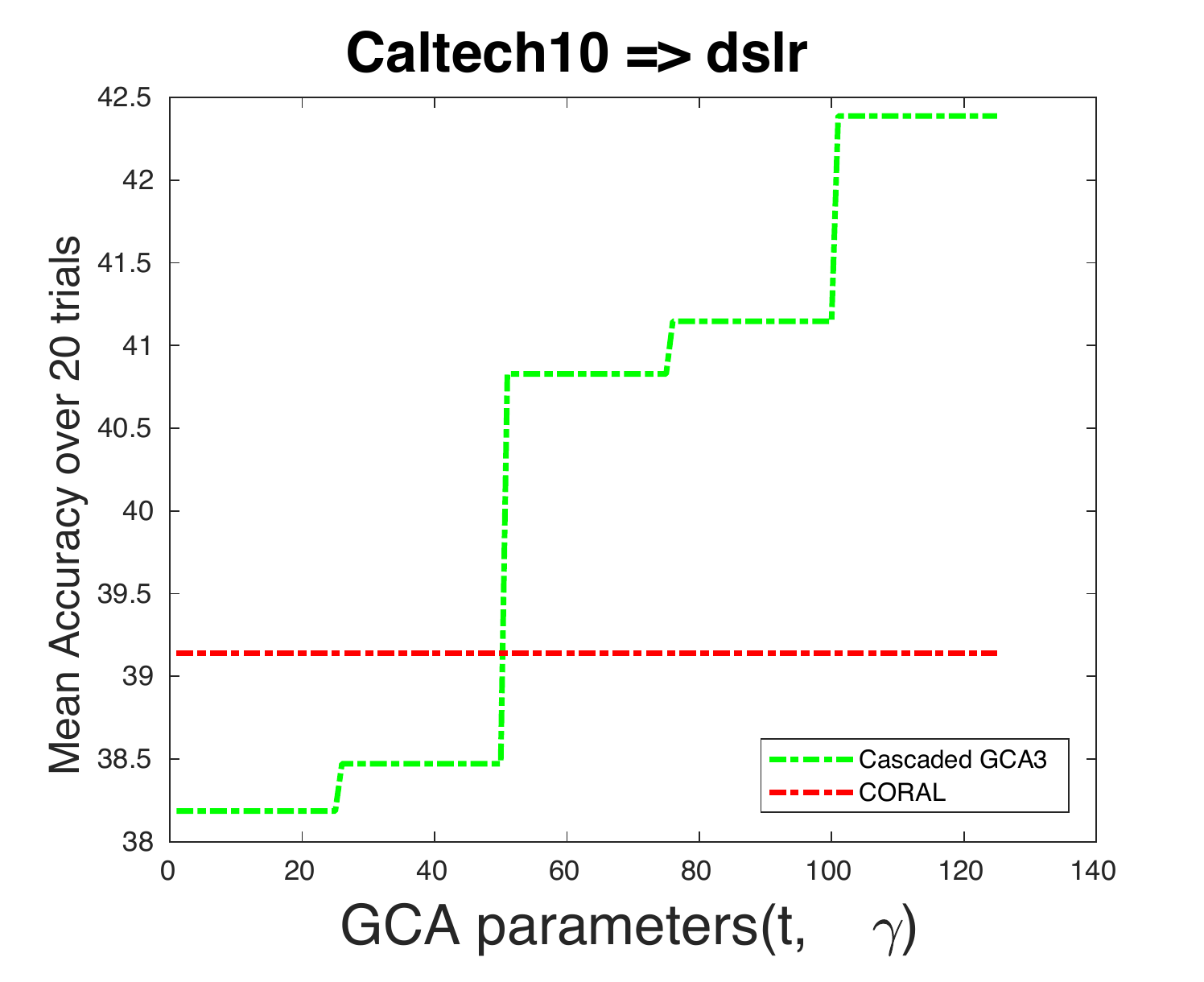}
	\caption{Comparison of the cascaded  GCA3 method with correlational alignment (CORAL). The parameters $t$ and $\gamma$ were varied between $0.1$ and $0.9$.}
	\label{fig:caltech10-to-dslr-gca3}
\end{figure}
\begin{figure}[p]
	\centering
	\includegraphics[width=0.8\linewidth]{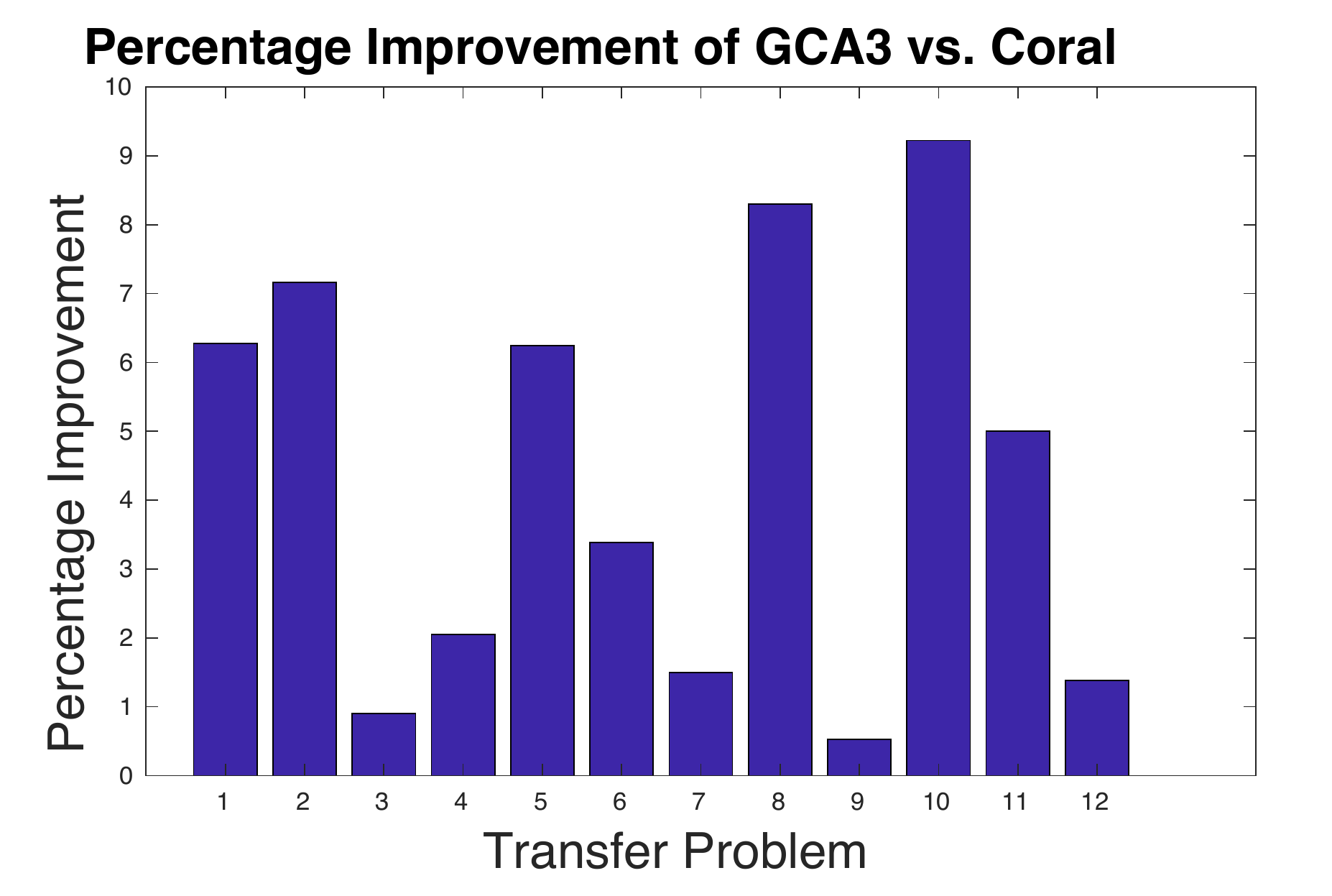}
	\caption{Comparison of the cascaded GCA3 method with correlational alignment (CORAL) in terms of percentage improvement.} 	
	\label{fig:gca3_perfimp}
\end{figure}

\begin{figure}[p]  
	\centering
	\includegraphics[width=0.8\linewidth]{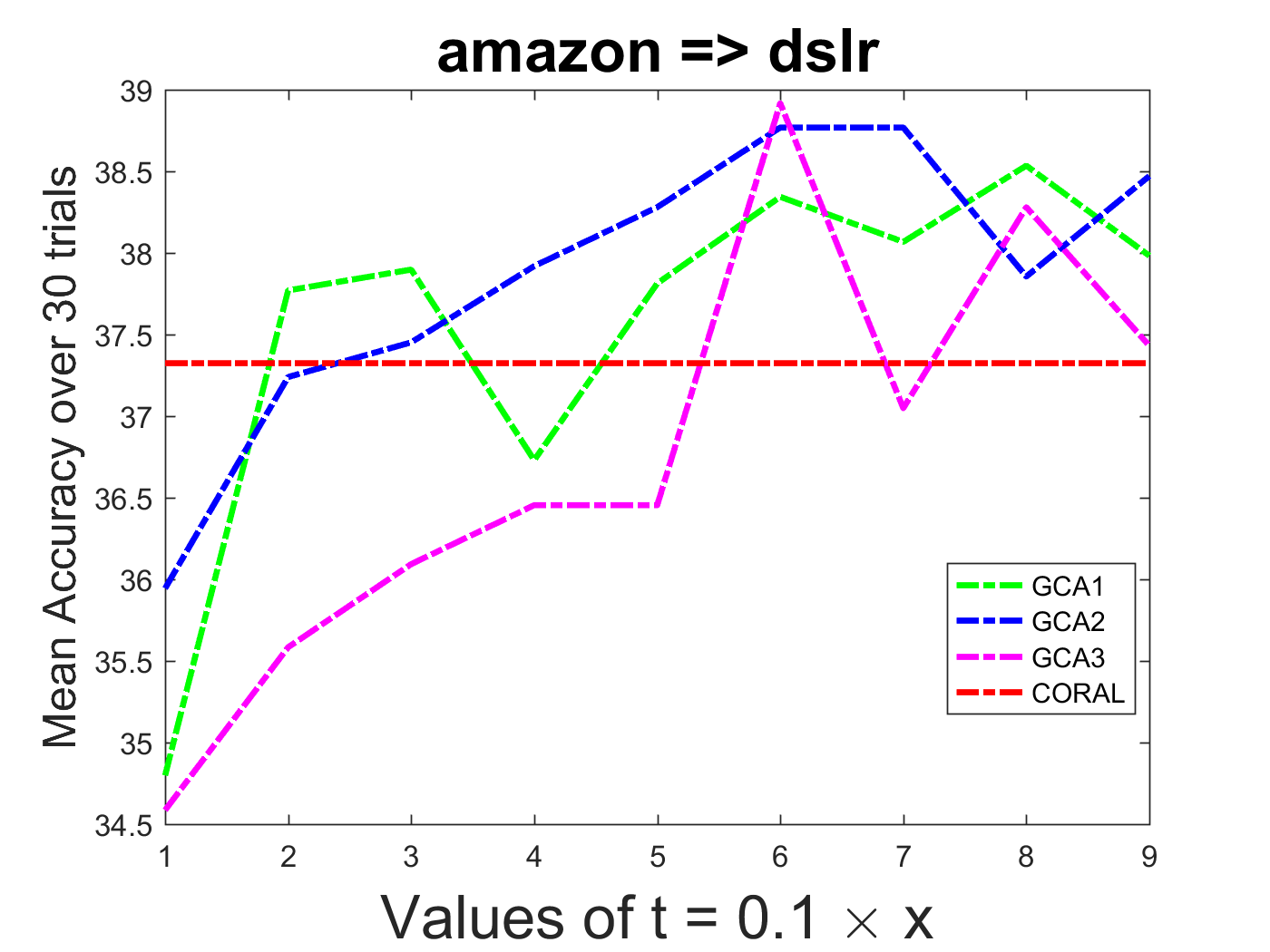}
	\caption{Comparison of the three proposed GCA methods (GCA1, GCA2, and GCA3) with correlational alignment (CORAL). }
	\label{fig:amazon-to-dslr-gca3}
\end{figure}

\begin{figure}[p]  
	\centering
	\includegraphics[width=0.8\linewidth]{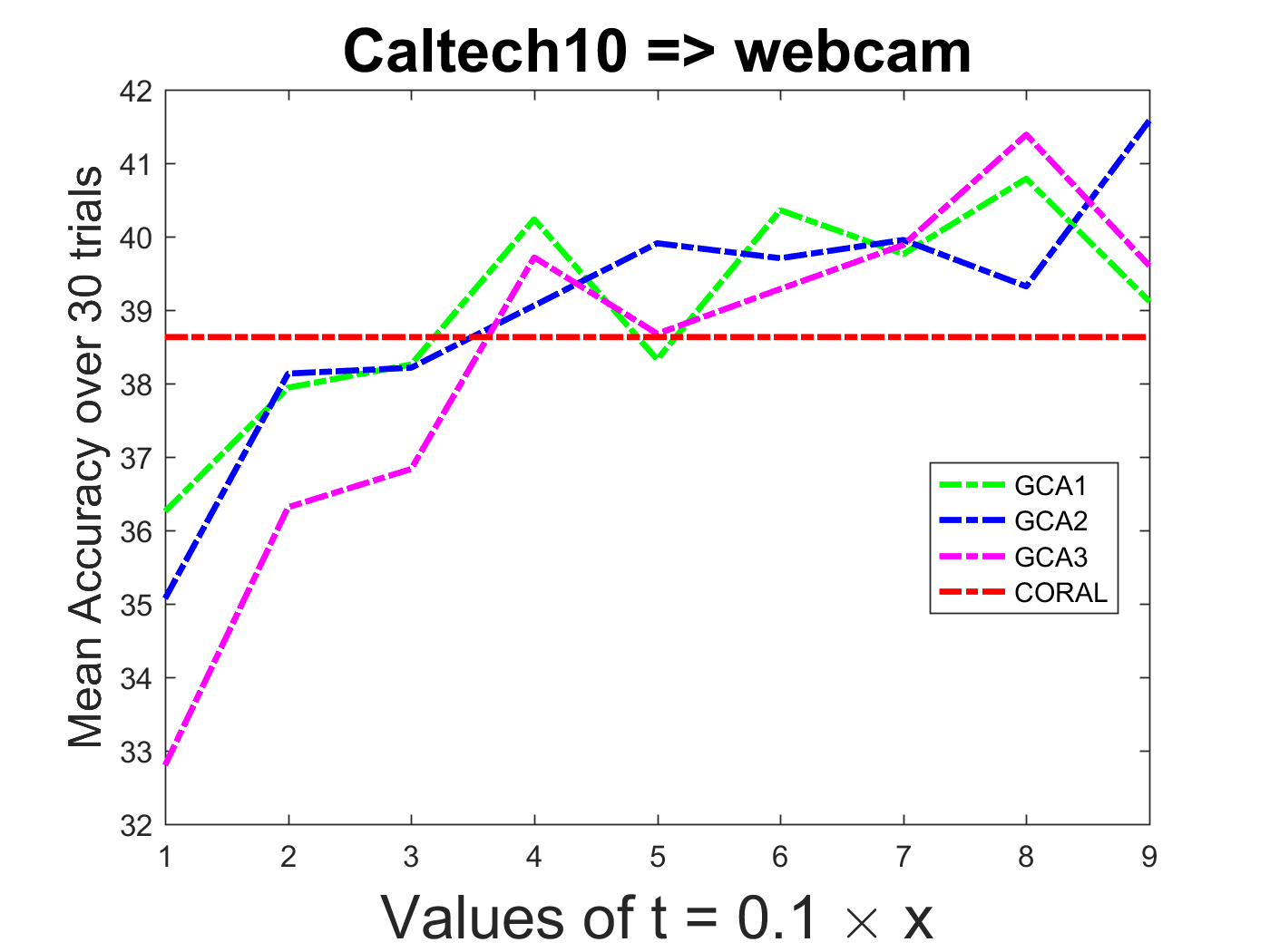}
	\caption{Comparison of the three proposed GCA methods (GCA1, GCA2, and GCA3) with correlational alignment (CORAL).}
	\label{fig:caltech10-to-webcam-gca3}
\end{figure}

One question that arises in proposed algorithms  is how to choose the value of $t$ in computing the weighted sharp mean. Figure~\ref{fig:caltech10-to-dslr-gca3} illustrates the variation in performance of the cascaded GCA3 method over CORAL over the range $t, \gamma \in (0.1, 0.9)$, and $\mu$ fixed for simplicity. Repeating such experiments over all 12 transfer learning tasks, Figure~\ref{fig:gca3_perfimp} shows the percentage improvement of the cascaded GCA3 method over correlational alignment (CORAL), using the best discovered value of all three hyperparameters using cross-validation. Figure~\ref{fig:amazon-to-dslr-gca3} compares the performance of the proposed GCA1, GCA2, and GCA3 methods where just the $t$ hyperparameter was varied between 0.1 and 0.9 for the {Amazon} to {DSLR} domain adaptation task. Note the variation in performance with $t$ occurs at different points for the three points, and while their performance is superior overall to CORAL, their relative performances at the maximum values are not very different from each other. Figure~\ref{fig:caltech10-to-webcam-gca3} once again repeats the same comparison for the {Caltech10} to the {Webcam} domain adaptation task. As these plots clearly reveal, the values of the hyperparameters has a crucial influence on the performance of all the proposed GCAXX methods. The plot compares the performance of GCA1 to the fixed performance of the CORAL method. 


%
%

\section{Summary and Future Work}

In this paper, we introduced a novel formulation of the classic domain adaptation problem in machine learning, based on computing the cascaded geometric mean of second order statistics from source and target domains to align them. Our approach builds on the nonlinear Riemannian geometry of the open cone of symmetric positive definite matrices (SPDs), using which the geometric mean lies along the shortest path geodesic that connects source and target covariances. Our approach has three key advantages over previous work: 
(a) Simplicity: The Riccati equation is a mathematically elegant solution to the domain adaptation problem, enabling integrating geometric and statistical information. 
(b) Theory: Our approach exploits  the Riemannian  geometry of SPD matrices. 
(c) Extensibility: As our algorithm development indicates, it is possible to easily extend our approach to capture more types of constraints, from geometrical to statistical. 

There are many directions for extending our work. Since we have shown how to reduce domain adaption to a problem involving metric learning, we can use other metric learning methods to design new domain adaption methods. This process can generate a wealth of new methods, some of which may outperform the method proposed in our paper. Also, while we did not explore nonlinear variants of our approach, it is possible to extend our approach to develop a deep learning version where the gradient of the three objective functions is used to tune the weights of a multi-layer neural network. As in the case of correlational alignment (CORAL), we anticipate that the deep learning variants may perform better due to the construction of improved features of the training data. The experimental results show that the performance improvement tends to be more significant in some cases than in others. A theoretical analysis of the reasons for this variance in performance would be valuable, which is lacking even for previous methods such as CORAL. \\


\section{Acknowledgments} 

Portions of this research were completed when the first author was at SRI International, Menlo Park, CA and when the second author was at Amazon.com, Bangalore, 560055, India. The first author currently holds a Visiting Professor appointment at Stanford, and is on academic leave from University of Massachusetts, Amherst.

\bibliographystyle{splncs04} 
\bibliography{daml-bib} 
\end{document}